\documentclass[11pt, a4paper]{article}
\usepackage[left=1.9cm, top=2.54cm, right=1.9cm, bottom=2.54cm]{geometry}
\usepackage[utf8]{inputenc}
\usepackage[english]{babel}
\usepackage{times}
\usepackage[T1]{fontenc}
\usepackage{mathptmx}
\usepackage{color}
\usepackage{amsmath}
\usepackage{amssymb}
\usepackage{amsfonts}
\usepackage{mathrsfs}
\usepackage{graphicx}
\usepackage{textcomp, gensymb}
\usepackage{hyperref}
\usepackage{tikz}
\usepackage{emptypage}
\usepackage{titlesec}
\usepackage{float}
\usepackage{amsmath}
\usepackage{algorithm}
\usepackage[noend]{algpseudocode}
\usepackage{subcaption}
\usepackage{graphicx}
\usepackage{comment}

\graphicspath{{img/}}

\newcommand\blfootnote[1]{
    \begingroup
    \renewcommand\thefootnote{}\footnote{#1}
    \addtocounter{footnote}{-1}
    \endgroup
}

\DeclareCaptionLabelFormat{andtable}{#1~#2  \&  \tablename~\thetable}

\usepackage{acro}
\DeclareAcronym{obu}{
	short			= OBU ,
	long			= On-board Unit,
}
\DeclareAcronym{rsu}{
    short           = RSU ,
    long            =Roadside Unit,
}

\DeclareAcronym{v2i}{
    short           = V2I,
    long            =Vehicle to Infrastructure,
}
\DeclareAcronym{i2v}{
    short           = I2V,
    long            =Infrastructure to Vehicle,
}
\DeclareAcronym{v2v}{
    short           = V2V,
    long            =Vehicle to Vehicle,
}
\DeclareAcronym{v2x}{
    short           = V2X,
    long            =Vehicle to Everything,
}

\DeclareAcronym{c-v2x}{
    short           = C-V2X,
    long            = Cellular Vehicle to Everything,
}
\DeclareAcronym{dcu}{
    short           = DCU,
    long            = Data-collection Unit,
}

\DeclareAcronym{vanet}{
    short           = VANET,
    long            =Vehicular Ad-Hoc Network,
}
\DeclareAcronym{manet}{
    short           = MANET,
    long            =Mobile Ad-Hoc Network,
}
\DeclareAcronym{tcp}{
    short           = TCP,
    long            =Transmission Control Protocol,
}
\DeclareAcronym{udp}{
    short           = UDP,
    long            =User Datagram Protocol,
}
\DeclareAcronym{wave}{
    short           = WAVE,
    long            =Wireless Access in Vehicular Environments,
}

\DeclareAcronym{lidar}{
    short           = LiDAR,
    long            =Light Detection And Ranging,
}
\DeclareAcronym{gps}{
    short           = GPS,
    long            =Global Positioning System,
}
\DeclareAcronym{denm}{
    short           = DENM,
    long            =Decentralized Environmental Notification Message,
}
\DeclareAcronym{cam}{
    short           = CAM,
    long            =Cooperative Awareness Message,
}
\DeclareAcronym{wifi}{
    short           = Wi-Fi,
    long            = IEEE 802.11 a/g/n,
}
\DeclareAcronym{au}{
    short           = AU,
    long            =Application Unit,
}
\DeclareAcronym{etsi}{
    short           = ETSI,
    long            =European Telecommunications Standards Institute,
}

\DeclareAcronym{rssi}{
    short           = RSSI,
    long            = Received Signal Strength Indication,
}
\DeclareAcronym{its}{
    short           = ITS,
    long            = Intelligent Transport System,
}
\DeclareAcronym{c-its}{
    short           = C-ITS,
    long            = Cooperative Intelligent Transport Systems,
}
\DeclareAcronym{its-s}{
    short           = ITS-S,
    long            = Intelligent Transport System Station,
}

\DeclareAcronym{nap}{
    short           = NAP,
    long            = Network Architectures and Protocols,
}
\DeclareAcronym{ieee}{
    short           = IEEE,
    long            = Institute of Electrical and Electronics Engineers,
}

\DeclareAcronym{iot}{
    short           = IoT,
    long            = Internet of Things,
}
\DeclareAcronym{iov}{
    short           = IoV,
    long            = Internet of Vehicles,
}
\DeclareAcronym{sdn}{
    short           = SDN,
    long            = Software Defined Network,
}

\DeclareAcronym{cpm}{
    short           = CPM,
    long            = Collective Perception Message,
}
\DeclareAcronym{cps}{
    short           = CPS,
    long            = Collective Perception Service,
}
\DeclareAcronym{dcc}{
    short           = DCC,
    long            = Decentralized Congestion Control,
}

\DeclareAcronym{mec}{
    short           = MEC,
    long            = Multi-access Edge Computing,
}
\DeclareAcronym{fc}{
    short           = FC,
    long            = fog computing,
}
\DeclareAcronym{vec}{
    short           = VEC,
    long            = Vehicular Edge Computing,
}

\DeclareAcronym{avr}{
    short           = AVR,
    long            = Augmented Vehicular Reality,
}
\DeclareAcronym{mpr}{
    short           = MPR,
    long            = Market Penetration Rate,
}
\DeclareAcronym{cav}{
    short           = CAV,
    long            = Connected and Autonomous Vehicles,
}
\DeclareAcronym{avod}{
    short           = AVOD,
    long            = Aggregate View Object Detection,
}

\DeclareAcronym{imu}{
    short           = IMU,
    long            =Inertial Measurement Unit,
}
\DeclareAcronym{vam}{
    short           = VAM,
    long            = Vulnerable Road User Awareness Message,
}
\DeclareAcronym{vru}{
    short           = VRU,
    long            = Vulnerable Road User,
}

\DeclareAcronym{spatem}{
    short           = SPATEM,
    long            = Signal Phase And Timing Extended Message,
}
\DeclareAcronym{mapem}{
    short           = MAPEM,
    long            = MAP (topology) Extended Message,
}
\DeclareAcronym{ivim}{
    short           = IVIM,
    long            = Infrastructure to Vehicle Information Message,
}
\DeclareAcronym{srem}{
    short           = SREM,
    long            = Signal Request Extended Message,
}
\DeclareAcronym{ssem}{
    short           = SSEM,
    long            = Signal request Status Extended Message,
}

\DeclareAcronym{dsrc}{
    short           = DSRC,
    long            = Dedicated Short-Range Communications,
}
\DeclareAcronym{lte-v2x}{
    short           = LTE-V2X,
    long            = Long Term Evolution Vehicle to Everything,
}
\DeclareAcronym{atcll}{
    short           = ATCLL,
    long            = Aveiro Tech City Living Lab,
}
\DeclareAcronym{gnss}{
    short           = GNSS,
    long            = Global Navigation Satellite System,
}
\DeclareAcronym{btp}{
    short           = BTP,
    long            = Basic Transport Protocol,
}
\DeclareAcronym{3gpp}{
    short           = 3GPP,
    long            = 3rd Generation Partnership Project,
}
\DeclareAcronym{lte}{
    short           = LTE,
    long            = Long Term Evolution,
}
\DeclareAcronym{nr-v2x}{
    short           = NR-V2X,
    long            = New radio Vehicle to Everything,
}

\DeclareAcronym{epm}{
    short           = EPM,
    long            = Environmental Perception Message,
}

\DeclareAcronym{mqtt}{
    short           = MQTT,
    long            =  Message Queuing Telemetry Transport,
}
\DeclareAcronym{json}{
    short           = JSON,
    long            = JSON,
}

\DeclareAcronym{ros}{
    short           = ROS,
    long            = Robot Operating System,
}

\DeclareAcronym{its-g5}{
    short           = ITS-G5,
    long            = Intelligent Transport Systems Generation 5
}

\DeclareAcronym{can}{
    short           = CAN,
    long            = Controller Area Network
}

\DeclareAcronym{lan}{
    short           = LAN,
    long            = LAN
}

\DeclareAcronym{vcu}{
    short           = VCU,
    long            = Vehicle Control Unit
}

\DeclareAcronym{agl}{
    short           = AGL,
    long            = Automotive Grade Linux
}

\DeclareAcronym{vcas}{
    short           = VCAS,
    long            = VRU Collision Avoidance System
}

\DeclareAcronym{sbc}{
    short           = SBC,
    long            = Single-board Computer
}

\DeclareAcronym{gpu}{
    short           = GPU,
    long            = Graphics Processing Unit
}

\DeclareAcronym{ap}{
    short           = AP,
    long            = Access Point
}

\DeclareAcronym{usb}{
    short           = USB,
    long            = USB
}

\DeclareAcronym{rtsp}{
    short           = RTSP,
    long            = Real Time Streaming Protocol
}

\DeclareAcronym{http}{
    short           = HTTP,
    long            = Hypertext Transfer Protocol
}

\DeclareAcronym{yolo}{
    short           = YOLO,
    long            = You Only Look Once
}
\DeclareAcronym{bsm}{
    short           = BSM,
    long            = Basic Safety Message
}
\DeclareAcronym{cdf}{
    short           = CDF,
    long            = Cumulative Distribution Function
}
\DeclareAcronym{rad}{
    short           = RAD,
    long            = Rolling Average Distance
}
\DeclareAcronym{ras}{
    short           = RAS,
    long            = Rolling Average Speed
}
\DeclareAcronym{pdf}{
    short           = PDF,
    long            = Probability Distribution Function
}
\DeclareAcronym{CV}{
    short           = CV,
    long            = Computer Vision
}
\DeclareAcronym{ml}{
    short           = ML,
    long            = Machine Learning
}

\titleformat{\section}{\normalfont\bfseries}{\thesection}{}{}
\titleformat{\subsection}{\normalfont\itshape}{\thesubsection}{}{}
\usepackage{parskip}
\usepackage[font=bf,labelfont=bf]{caption}
\usepackage{fancyhdr}
\begin{document}\thispagestyle{empty}

%%%%%%%%%%%%%%%%%%%%%%%%%%%%%%%%%%
%%%% Não alterar, cabeçalho do Template

\begin{tikzpicture}[remember picture, overlay]
\node(Logo) at (current page.north west) [anchor=north west, xshift=1.8cm,yshift=-0.7cm]{\includegraphics[width=3.51cm, height=2.72cm]{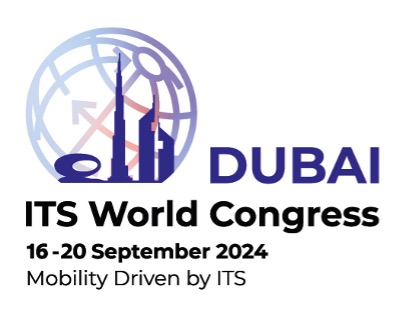}};
\node at  (7.2,-1) {30th ITS World Congress, Dubai, UAE, 16-20 September 2024};
\end{tikzpicture}
%%%%%%%%%%%%%%%%%%%%%%%%%%%%%%%%%%
    \begin{center}
    \vspace{1cm}
    % Paper ID
    {\fontsize{14pt}{\baselineskip}\selectfont Paper ID 333 } \\
    \vspace{0cm}
    % Título
    {\fontsize{14pt}{\baselineskip}\selectfont \textbf{Real-time Object and Event Detection Service through Computer Vision and Edge Computing}}
    \vspace{0cm}
    
    % Nome dos autores
    \textbf{Marcos Mendes$^1$, Gonçalo Perna$^1$, Pedro Rito$^1$, Duarte Raposo$^1$ and Susana Sargento$^{1,2}$}\\
    {\fontsize{10pt}{\baselineskip}\selectfont $^1$Instituto de Telecomunicações, 3810-193 Aveiro, Portugal\\
    $^2$Departamento de Eletrónica, Telecomunicações e Informática, Universidade de Aveiro, 3810-193 Aveiro, Portugal}
    \vspace{0cm}
    
    % Resumo
    \end{center}
    
    \textbf{Abstract}\\
    \vspace{-0.7cm}
    %Road accidents are estimated to cost hundreds of billions of dollars annually. Figures from the World Health Organization suggest that road traffic crashes cost approximately 518 billion dollars globally each year, which accounts for 3\% of the gross domestic product for most countries.The road safety situation in the European Union (EU) showed increased road deaths in 2021 compared to 2020, but a decrease compared to 2019 pre-pandemic. The average road deaths per million inhabitants in the EU in 2021 was 44, while in Portugal, it was 52.

    %Most fatal road accidents in urban areas involve \acp{vru}. New solutions for fighting against these accidents can be considered using advanced technology, leveraging sensors, large datasets, \ac{ml} models, communication and edge computing.
    %This paper presents a strategy and an implementation of a system for road monitoring and safety for Smart Cities, based on \ac{CV} and edge computing. % and links many areas of knowledge, modular, scalable and can be applied in other contexts.
    %Promising results were obtained by implementing vision algorithms and tracking in a surveillance camera, part of a Smart City testbed, the \ac{atcll}. The algorithm accurately detected cars, pedestrians, and bicycles while predicting road state and inferring on collision events to prevent collisions, in almost real-time. %{The tests done to validate this work its performed with the \ac{atcll} infrastructure. 
    
    The World Health Organization suggests that road traffic crashes cost approximately 518 billion dollars globally each year, which accounts for 3\% of the gross domestic product for most countries. Most fatal road accidents in urban areas involve \acp{vru}. Smart cities environments present innovative approaches to combat accidents involving cutting-edge technologies, that include advanced sensors, extensive datasets, \ac{ml} models, communication systems, and edge computing. This paper proposes a strategy and an implementation of a system for road monitoring and safety for smart cities, based on \ac{CV} and edge computing. Promising results were obtained by implementing vision algorithms and tracking using surveillance cameras, that are part of a Smart City testbed, the \ac{atcll}. The algorithm accurately detects and tracks cars, pedestrians, and bicycles, while predicting the road state, the distance between moving objects, and inferring on collision events to prevent collisions, in near real-time.\blfootnote{This work was supported in part by the EU’s HE research and innovation programme HORIZON-JU-SNS-2023 under the 6G-PATH project (Grant No. 101139172). The views expressed in this contribution are those of the authors and do not necessarily represent the project nor the Commission.}

    \vspace{-0.3cm}
    
    % Palavras-chave
    \noindent
    \textbf{Keywords: } \\ \MakeUppercase{Object Tracking, Collision Avoidance, Smart City, Multi-access Edge Computing.}
    %\vspace{1cm}
    
% Body 
\noindent

\section{Introduction}
%The socioeconomic impact of road accidents extends beyond the monetary aspect. It touches upon issues of public health, safety regulations, infrastructure development, and the need for effective road safety measures and policies. 

Road accidents are estimated to cost hundreds of billions of dollars annually, according to the World Health Organization~\cite{Roadtraf5:online}. These numbers suggest that road traffic crashes cost approximately 3\% of the gross domestic product for most countries~\cite{Roadtraf5:online}. The road safety situation in the European Union (EU) showed increased road deaths in 2021 compared to 2020, but a decrease compared to 2019 pre-pandemic. The average road deaths per million inhabitants in the EU in 2021 was 44~\cite{Mobility_and_Transport}. Governments, international organizations, and various stakeholders recognize the urgency of addressing road safety concerns. The European Union's Road Accident Statistics Database serves as a comprehensive source of information regarding road fatalities, specifically focusing on \acp{vru}, in European Union Member States and other countries. The Road Accident Statistics Database provides detailed insights into various factors associated with road fatalities, including age, gender, type of road user, time of day, road surface, alcohol and drug use, type of collision, and speed~\cite{Mobility_and_Transport}. 

\ac{CV}-based accident detection systems utilize image processing and \ac{ml} algorithms to detect and classify traffic accidents in real-time. These systems are capable of identifying and categorizing various objects such as cars, pedestrians, and bicycles, enabling further analysis and intervention~\cite{Computer_Vision-based_Accident}. The core component of these systems is the \ac{CV} algorithm, which analyzes captured images performing tasks such as object recognition, motion detection, and position estimation~\cite{Computer_Vision-based_Accident}. Overall, \ac{CV}-based accident detection systems offer a promising solution to improve road safety and response efficiency. Their ability to analyze real-time video data, detect objects, and predict potential incidents can have a significant impact on accident prevention and mitigation efforts. However, for a depth, or distance measurement, a two camera system normally needs to be implemented. This is an expensive solution and cannot be implemented in every case. In this work it is proposed a new method for a single camera to measure distances and be able to accurately detect collision risks in the roads. This approach is specifically addressing static cameras.

%The presented research builds upon the principles of Computer Vision (CV), such as, object detection, classification, and depth perception, all performed through a single static camera. Throughout this study, various metrics, including latency, object detection, and their reliability, are meticulously considered. The primary goal is to characterize urban traffic congestion~\cite{posts}, allowing for the assessment of congestion levels at different times of the day. Parameters related to driving safety, such as speed, acceleration, and braking distance, were also thoroughly examined.

The proposed approach builds upon the principles of \ac{CV}, such as, object detection, classification, and depth perception, all performed through a single static camera. The developed work joins object detection and geographical location to determine the accurate position of detected moving objects, their distance and the probability of collision. This makes possible to analyse, through a video stream, new object information such as direction, velocity, acceleration and braking distance, which is not available in the related work when considering only a single camera. This work provided insight on the developed algorithm precision in detection and distance measuring, thus demonstrating the possibility of assessing road state, through a low cost implementation and reduced latency. The results show that the accuracy of the approach is good, and can be even increased if the definition of the video can be improved.

%This work also has in consideration metrics for road safety, and real-time analysis of road state, such as \ac{ras} and \ac{rad}.

The structure of this paper is outlined as follows. The "Related Work" section delves into existing literature related to the topic under discussion. Subsequently, the "Architecture and Implementation" section provides an overview of the overall system and details the implementation of the necessary services. Following this, the "Services Performance" section expounds on the results of object detection, including the validation of tracking and distance measurements. Moving forward, the "Collision Detection and Risk" section introduces the algorithm used to calculate the probability of collision between two objects, along with its real detection results. Lastly, the concluding section summarizes the key findings of the study and outlines potential avenues for future work.
\section{Related Work}
    Considering the proposed goals of this paper, it is important to consider related work such as use cases regarding mobility, detection algorithms and camera processing features. This section also considers other works for mobility focusing in interactions of \acp{vru} and other road users, since it is important to understand how the road users can interact.

%The authors in~\cite{inteligent_traffic_tool} developed a system for counting and monitoring the speed of vehicles in a toll system, where road laws should not be violated and monitoring is constantly needed%, but this implementation cannot be instantly deployed meaning configuration in each point is needed
To understand \acp{vru} mobility is essential to find a way for tracking and identifying each one individually. In~\cite{heat_santander} a study was performed in a market place, where WiFi \ac{iot} sensors and two cameras were installed and two pilot studies were made to evaluate eventual choke points in the market. The work in~\cite{sensor_based} presents an approach with internet node base to calculate the number of users linked to the same WiFi. This tracking is based on the MAC address of each individual user. This is a great approach for general position acquisition, but does not take in consideration users with more than one connected device, and users with no devices connected, being able to understand stress over the network, but cannot be considered as a viable application to a counting and tracking problem. As accidents occur, only knowing \ac{vru} general position has little to no effect, since cars cannot stop or alert before colliding. 

%Newer vehicles now come equipped with many \ac{iot} sensors. In~\cite{adas}, it is considered the possibility to regulate the speed by having either fixed sensors along the highway, or a GPS solution, only to regulate vehicle speed based on these metrics, lacking \ac{CV} to analyse the road usage and state around the vehicle. The implementation of vehicle Advanced Driver Assistance Systems (ADAS) is widely documented as examples of lane detection~\cite{lane_tracking}. Knowing the lanes position makes possible to determine where the car is relative to the road. This allows small angle corrections to the steering wheel, using the help of Hough Transforms\footnote{https://www.analyticsvidhya.com/blog/2022/06/a-complete-guide-on-hough-transform/}. As this system is implemented, it lacks the possibility to detect if other vehicles are in front of the vehicle, meaning that it would not prevent a collision by following the lane, not checking for the need to break. These two solutions never take in consideration the car surroundings, and if a user diverges from what is expected, traffic accidents may occur.

To assess potential risk situations, it is crucial to comprehend the movement patterns of \acp{vru}. In \cite{angle}, the flow of the road is evaluated based on the first and last detected angles of movement. Unlike focusing on collision detection, this research investigates the relevance of a vehicle's turning angle to its trajectory. This assessment is conducted using a vectorial coordinate system specific to the vehicle, without considering global positioning. The efficient monitoring of traffic is integral to a well-functioning city. Typically, this responsibility falls on traffic police; however, their presence everywhere is impractical for preventing and resolving traffic issues. With the advancements in Machine Learning (\ac{ml}), vision algorithms have significantly improved. These algorithms are now employed in analyzing traffic conditions and utilizing traffic lights as a management tool~\cite{yolov7_traffic}. Real-time feeds are utilized to assess the current state of the road~\cite{real_time_traffic_yolov7}, marking a significant leap forward in the capabilities of traffic analysis and management systems.

Use cases such as the mentioned in~\cite{crosswalks_usecase}, acknowledge the possibility of implementing \ac{ml} with traffic management software, resulting in efficient mobility. This work presents two new topics for mobility: distance measure with only one static camera, and risk assessment based on distance measured.
Considering the conclusion on~\cite{decisions}, pedestrian and traffic sign detections provide vital information and a lot of difficulties, making challenging for \ac{ml} algorithms to evaluate the surroundings in different conditions compared to the ones they are trained on. \ac{CV} algorithms can be implemented and provide accurate classification for road traffic situations such as \ac{yolo}, R-CNN, F-CNN~\cite{traffic_evaluation}.

%It is important to recognize, accurately, different types of \acp{vru}. This can be accomplished using the \ac{yolo} algorithm. There are several versions, all with different characteristics, of the object detection algorithm with important metrics such as the ones represented in Table~\ref{tab: yolomodels}. The mean average precision provides an overall assessment of the algorithm's performance, taking into account the accuracy across different classes. Higher mean average precision values indicate better performance. These versions differ in the size of the input image they can handle, the number of classes they can recognize, and the speed at which they can process images. Newer versions of the YOLO model have been developed currently. YOLOv8 model is not included in Table~\ref{tab: yolomodels} because it can be considered a mix between YOLOv3 and YOLOv5, and its performance can only be distinct due to the hardware conditions.

\begin{table}
    \centering
    \begin{tabular}{|c|c|c|c|c|}
        \hline
        \textbf{Model} & \textbf{Input size} & \textbf{Classes} & \textbf{mAP} & \textbf{fps} \\ \hline
        YOLOv1& 448x448 & 20 & 63.4 & 45 \\ \hline
        YOLOv2 & 448x448 & 20 & 57.9 & 78 \\ \hline
        YOLOv3 & 608x608 & 80 & 57.9 & 22 \\ \hline
        YOLOv3-tiny & 416x416 & 80 & 44.7 & 65 \\ \hline
        YOLOv4 & 640x640 & 100 & * & \# \\ \hline
        YOLOv4-tiny & 416x416 & 100 & * & \#  \\ \hline
        YOLOv5s & 640x640 & 100 & * & \#  \\ \hline
        YOLOv5m & 640x640 & 100 & * & \#  \\ \hline
        YOLOv5l & 640x640 & 100 & * & \#  \\ \hline
        YOLOv5x & 640x640 & 100 & * & \#  \\ \hline
        YOLOv5xx & 640x640 & 100 & * & \#  \\ \hline
        YOLOv5-tiny & 640x640 & 100 & * & \#  \\ \hline
    \end{tabular}
    \caption{YOLO models basic characteristics comparison.}
    \label{tab: yolomodels}
    \vspace{1ex}

     {\raggedright * - Values for mean average precision are not mentioned because the developers of the model have chosen not to report them. This could be for a variety of reasons, such as the model was not designed to maximize mean average precision, or because the model is intended to be used for a specific task or application where mean average precision is not a relevant metric. \par
     
     \raggedright \# - The fps value, not specified in the table, will depend on the hardware it is running on, as well as the specific architecture and configuration of the model. However, these values may vary based on the specific implementation and hardware configuration. }
    \label{table:yoloall}
\end{table}

%As a city grows, it is important for the the technology to scale with the city. This will mean that the system will not have to be changed or customized for each deployment, except for a configuration file, but will need to be cheap and provide safety to the citizens. As described in~\cite{price_cameras}, surveillance cameras are the preferable surveillance sensor as they are versatile, easy to deploy and can enhance public safety. With a single camera implementation, it would not be possible to have a depth or a real relative distance perception, but in~\cite{camera_coordenates}, it is given the mathematical properties to correspond geographical coordinates to frame pixels and with only one camera. By using this approach in our solution, it is possible to obtain real geographical position of detected objects, with relatively good accuracy.

The use of \ac{CV} can raise concerns about privacy, particularly when it comes to the collection and use of personal data. Systems of this nature often collect large amounts of data, which could be used to build detailed profiles of individuals and their activities~\cite{intro_privacy}. To balance the need for security with the protection of user privacy, many companies and organizations have implemented strict guidelines and policies for the use of \ac{CV} technology \cite{blur_redact}. The General Data Protection Regulation in the European Union applies specific provisions related to the protection of personal data used for automated decision-making, including \ac{CV} systems \cite{GDPR}. When applying \ac{CV}, it is essential to consider three types of privacy: data privacy, model privacy, and deployment privacy \cite{Privacy_cv}.

%\textcolor{red}{PR: falta discussão}
The aim of this work is to perceive and measure the distance between \acp{vru} and other road users using a state-of-the-art detection algorithm and a single camera that can estimate the global position accurately. The proposed algorithm also considers the vehicle trajectory and road safety metrics, and offering a scalable and low-cost solution for real-time road assessment.
\section{Architecture and Implementation}

The objective of this work is to ensure and improve mobility safety in a city environment, allowing for road state evaluation and risk assessment inside specific scenarios, all in real-time. This will provide enough information to possibly predict if an accident is eminent of happening. This will be performed through the detection and tracking of moving objects using video cameras and processing on the edge of the network. The implementation of vision algorithms in smart cities involves the integration of various technologies to improve the quality of life for urban residents. To make the implementation more efficient and usable for scenarios requiring low-latency, we consider the usage of a single camera. %, as it becomes more affordable and easy to implement, covering a wider area with fewer video cameras. %and having already that implementation characteristics all over the city of Aveiro. This will allow the algorithm to run in already implemented posts, but with customized configurations for each station, since GPS is different for each post.

To consider implementing this work on a city platform it is needed a system that has the capability of capturing and processing the information. Firstly, we have to consider a infrastructure with video cameras spread throughout the city and close to areas of interest, such as road intersections. Attached to the video cameras, the infrastructure includes edge computing capabilities (for the object detection and algorithm processing) with backhaul communication to the cloud (for management of the system). In order to deploy the services, the system requires few preparation steps to guarantee the accuracy of the detection.

%As this work is divided in three services each one has different requirements as the first, responsible for the frame acquisition, 1) a tool enabling video streaming 2) a process for calibrating the camera 3) an object detection service 4) a position calculation service and finally, 5) a data sending and processing service for interfacing with other services. For the first GStreamer was implemented, for the third \ac{ml} model was used and for the fifth an MQTT broker was used.

\subsection{System Preparation}

The preparation of the solution is comprised by two main steps, which are required to be set in every edge site: (1) camera calibration, to compensate the camera lens distortion, ultimately improving the detection and the classification, and (2) geo-framing calibration, so that a matrix can be pre-loaded, and then for every position of the frame it is possible to get an approximate GPS position of the object.

The first step, calibrating the camera, means the process of extracting the lens distortion coefficients. As these coefficients are obtained, this makes possible to erase the distortion provoked by the lens. In this work calibration was preformed with chessboard pattern which is a common method employed in computer vision and computer graphics to determine the intrinsic parameters of a camera~\cite{calibration}. The intrinsic parameters include the camera's focal length, principal point, and lens distortion coefficients. These parameters are crucial for correcting distortions and obtaining accurate measurements in images.

The global positioning calculation is done by selecting four non-collinear points on the frame, preferably on the road, and obtaining each point's correspondent geographic coordinate, in a process called geo-framing~\cite{camera_coordenates}. The accuracy of the taken points and their correspondent GPS coordinates ensure the system accuracy. An important note is that this method assumes a broad view of the ground within the frame, so that the coordinate mesh turns out to be more accurate. % where the ground is represented, in the frame, this is important to consider, since center of detection does not represent the position and it needs to shift to the detection nearest ground point.

\subsection{Services and Deployment}

\begin{figure}[ht]

  \centering
  \includegraphics[width= 0.8\linewidth]{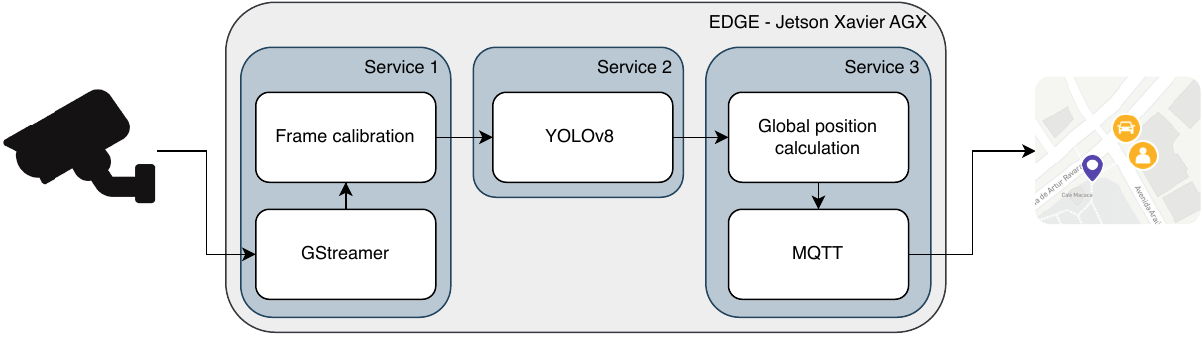}
  \caption{System architecture diagram.}
  \label{fig: diagram sol}
\end{figure}

The proposed architecture has three services concurrently running, since they have a different purpose: (1) the frame acquisition and then the compensation for the lens distortion (through the matrix multiplication obtained in the calibration); (2) the application of the detection algorithm; and (3) the calculation of the global geographical position. This architecture, presented in Figure~\ref{fig: diagram sol}, has a modular implementation and enables the seamless deployment and sequential execution.

% The implemented system is defined by three services, all operating on the edge. The first service takes requests, through GStreamer using the \ac{rtsp} protocol, and calibrates the frame, thus preparing it for the next service. The second service is the one where the image is processed and classifications are obtained. The third service, based on a pre-generated LookUp Table~\cite{camera_coordenates}, to map each pixel in the frame to real geographic coordinates. It publishes the detection results to an MQTT Broker.

%To deploy solution, several considerations are important to asses, such as the stream viability, the frame distortion of the camera lens, the detection algorithm and all the configurations and adaptability needed for the better performance of the model prediction, such as the hyper-parameters and finally camera and stream latency, to check for bottlenecks in the architecture.

%The system architecture is presented in Figure~\ref{fig: diagram sol}. 

To evaluate the best way for the frame acquisition, two solutions are considered: the use of \ac{rtsp} directly or through the GStreamer\footnote{\url{https://gstreamer.freedesktop.org/}} pipeline. In the end, the GStreamer was chosen due to its performance in providing a more fluid video stream. The operation of removing distortion is applied to the stream, by a simple matrix multiplication, every time a new frame is received. These two tasks complete the Service 1 of the architecture.

To detect and track \acp{vru}, it is important to choose a diversified and state of the art model, so that it can detect the \acp{vru} and other road users (vehicles, bikes, trucks, etc) and classify them correctly. For this approach, \textit{YOLOv8n}\footnote{\url{https://github.com/ultralytics/ultralytics}} is the chosen model for the detection Service 2. This choice is based on the comparision made in the previous section, considering YOLOv8 is based on YOLOv5. However, YOLOv8 is faster and more accurate than YOLOv5, and it provides a unified framework for performing object detection, instance segmentation, and image classification.

Finally, the last service calculates the GPS coordinate multiplying the pixel positions by the geo-framing matrix obtained in the preparation and the DISTANCECALCULATION function outputs a distance between the objects. Thereafter, the metadata of the detected objects is published to the MQTT broker, in a relevant topic, for other services to subscribe.

These services are all deployed in the edge of the network, located in the road side infrastructure, specifically in Jetson Xavier AGX due to its GPU capabilities. It is also important to mention that each edge node can process more than one frame gathering service (Service 1). This helps the management of the edge nodes in the infrastructure to distribute the processes. If an edge node becomes limited in resources or unavailable, due to hardware limitations, the services are migrated to the nearest and available edge node. The Jetson Xavier AGX is equipped with the Jetpack version \textit{4.6.4-b39}. % and runs the YOLOv8n model. The new YOLOv8 model allows the detection and identification as close as possible to the state of the art, and therefore, it will be used as is in our approach. %This article focus is the implementation of new findings taking in consideration previous work done.

\section{Services Performance}

In this section an analysis of services is performed, accounting for the end-to-end latency of the streaming and object detection services and its validation, and, finally, a test was also performed to confirm the measured distances.

%To evaluate risk situations, it is important to track and calculate distances between \acp{vru} and vehicles. Based on the related work and the proposed architecture, it is now possible to make such calculations. This section describes the different steps to perform the tracking and determine the distances between objects in the road, complementing with the obtained results.

%\subsection{Integration in the ATCLL platform}

The system was deployed in a real city-scale testbed and infrastructure in Aveiro, Portugal. The \ac{atcll}~\cite{actll}, deployed by \textit{Instituto de Telecomunicações} with the help of the Municipality of Aveiro, contains 44 access points (most in the form of smart lamp posts) throughout the city connected through fiber, with communication, edge computing and sensing devices. %, located as shown in Figure~\ref{fig: aveiro post location}. The blue arrows represent the smart lamp posts used in this work, which contain video cameras, radars and edge computing devices, the Jetsons Xavier AGX.}

\begin{comment}

\begin{figure}[ht]
    \centering
    {\includegraphics[width= 0.3 \linewidth]{img/aveiro slp.png}}
    \caption{Map of Smart Lamp Post used in Aveiro, from \ac{atcll}}
    \label{fig: aveiro post location}%
\end{figure}

\end{comment}

\subsection{Latency of Video detection}
%With the architecture and algorithms specification, it is required to analyse the total latency of the detection and processing pipeline.% the complexity and the accuracy of the proposed approach. This section will focus on that and provide a explanation on the service. } %This will be made by simplifying the explanation of each service and analyse some taken measures. 

\begin{figure}[H]
    \centering
    \includegraphics[width = 0.7\linewidth]{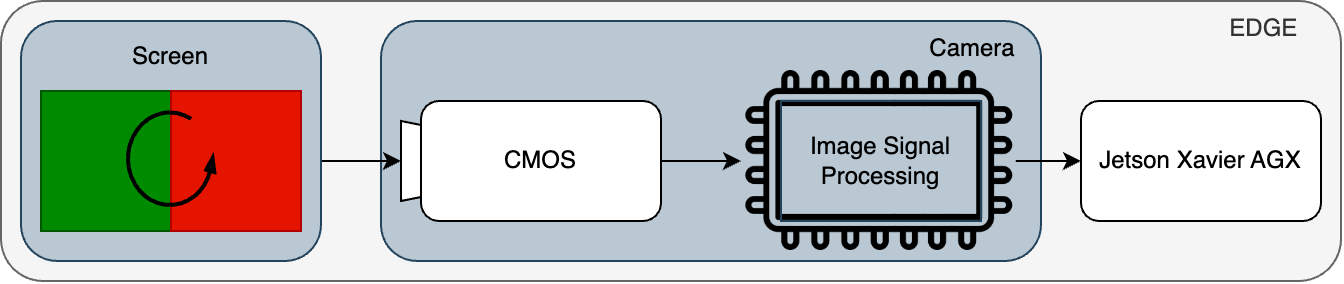}
    \caption{System latency measurement diagram.}
    \label{fig: lat diagram}
\end{figure}

This setup is designed to measure the latency, which starts from the time instant the video camera captures the frame to the moment it is displayed on a computer screen, as illustrated in Figure~\ref{fig: lat diagram}. As shown in the Figure, this consists on the change of colors, in an external monitor. As the monitor color changes from green to red, timestamps are taken. As the camera has a delay associated with the processing of each frame, it is important to include that delay in the frame acquisition time, this includes the light capture made by the CMOS sensor and the needed processing made by the image signal processing, taking also in consideration the time it takes for the camera to send the frame to the Jetson. This test ran continuously for a duration of 16 min. 

\begin{figure}[ht]
    \centering
    \begin{minipage}[b]{0.45\linewidth}
        \centering
        \includegraphics[width=\linewidth]{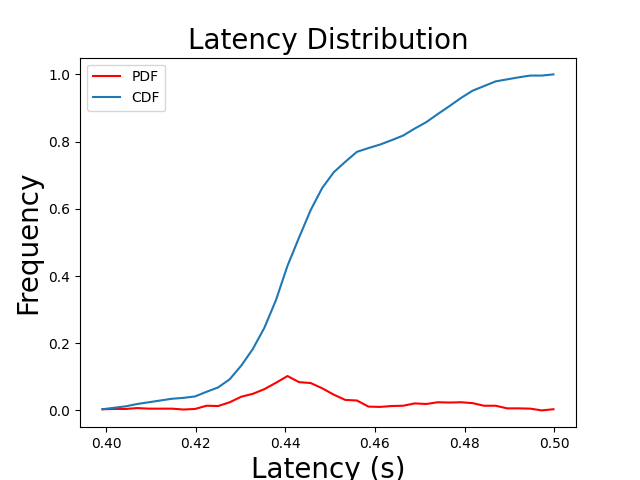}
        \caption{System latency \ac{cdf} and \ac{pdf}.}
        \label{fig: latency cdf}
    \end{minipage}
    \hspace{0cm}
    \begin{minipage}[b]{0.45\linewidth}
        \centering
        \includegraphics[width=\linewidth]{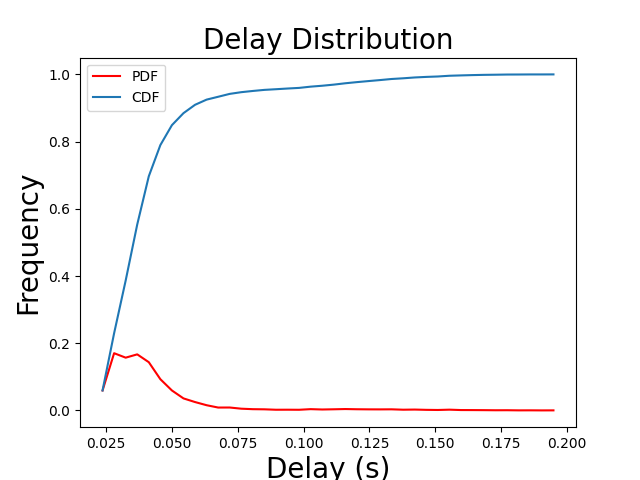}
        \caption{GStreamer latency \ac{cdf} and \ac{pdf}.}
        \label{fig: Server Delay CDF}
    \end{minipage}
\end{figure}

The obtained delay is shown as a \ac{cdf}, representing that the delay will take a value lower than or equal to the chosen value, and the \ac{pdf}, representing the immediate probability that a certain value of delay is measured. For the precise measurement, a video camera is positioned towards a monitor, using color changes on the monitor as reference points for timing. This approach reveals that the total service median latency is $0.44$ seconds, as indicated in Figure~\ref{fig: latency cdf}, with a minimum value of 0.35s, and a maximum value of 0.99s (standard deviation of 0.05s). 

As described before, the system contains the Jetson with GStreamer. This service is also assessed in terms of its latency, in Figure~\ref{fig: Server Delay CDF}. The mean latency is 0.08 s, and the standard deviation is 0.01 s. These tests were conducted with a fiber optics connection to the edge containing the system.

To benchmark the latency of the system, we refer to the standards set by the \ac{etsi} and compare them with the maximum latency specified in safety use cases for road users warnings, which is set at $300ms$~\cite{etsi_time}. %While this aligns with our expectations, aiming for real-time assessment pushes the boundaries of computer processing. 
The analysis reveals that the primary bottleneck contributing to latency lies in the camera's image acquisition, which is a fixable solution as the market provides faster cameras.

\subsection{Object Detection}

\begin{figure}[ht]
    \centering
    \begin{subfigure}{0.45\linewidth}
        \includegraphics[width=\linewidth]{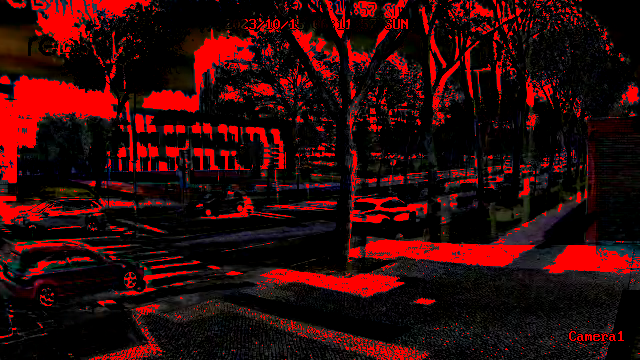}
        \caption{Difference between distorted and processed frames.}
        \label{fig: frame diff}
    \end{subfigure}
    \hfill
    \begin{subfigure}{0.45\linewidth}
        \includegraphics[width=\linewidth]{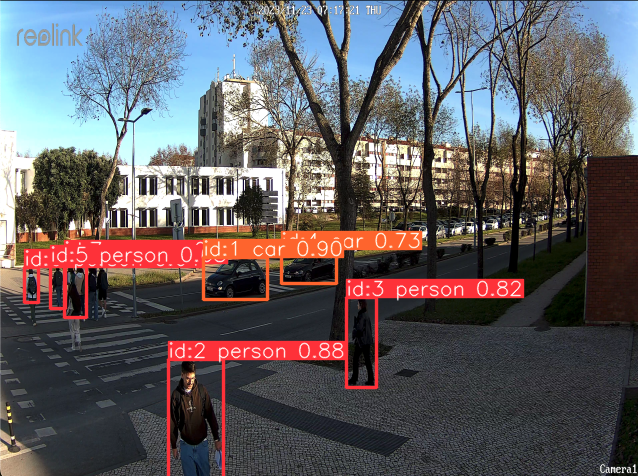}
        \caption{Frame with object identifiers.}
        \label{fig:detection_frame}
    \end{subfigure}
    \caption{Frame calibration and tracking object.}
\end{figure}

As discussed in the previous section, \textit{YOLOv8n} is used to detect and classify all objects in the road, in this case, pedestrians, motorbikes and bicycles, as showed in Figure~\ref{fig:detection_frame}. Using this method, data from each individual object can be obtained, and data manipulation can take place to obtain more information, such as velocity and acceleration. In this work, it is important to have a good perception of the real world, via the video camera, so that all validations can be as most accurate as possible. In Figure~\ref{fig: frame diff} it is shown the differences between the distorted frame and the processed frame, without the lens distortion.

% \begin{figure}[h]
%     \centering
%     \includegraphics[width=0.8 \linewidth]{img/distort.png}
%     \caption{Lens distortion matrix applied to frame.}
%     \label{fig: frame calib}
% \end{figure}

\begin{figure}[h]
    \centering
        \begin{subfigure}{0.45\linewidth}
        \includegraphics[width=\linewidth]{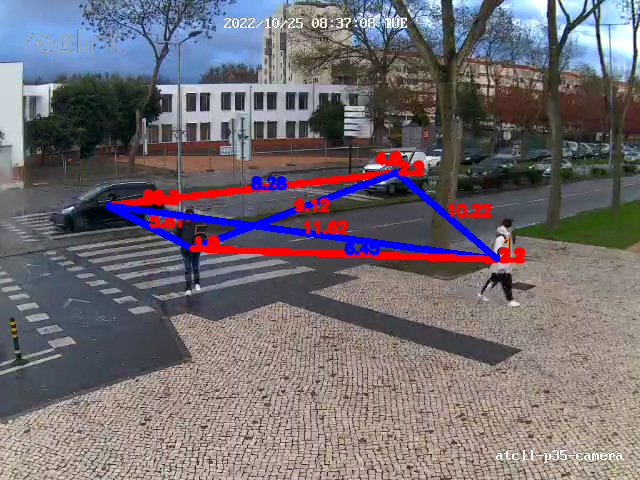}
        \caption{Distances between different objects.}
        \label{fig:calculation_frame}
    \end{subfigure}
    \hfill
    \begin{subfigure}{0.45\linewidth}
        \includegraphics[width=\linewidth]{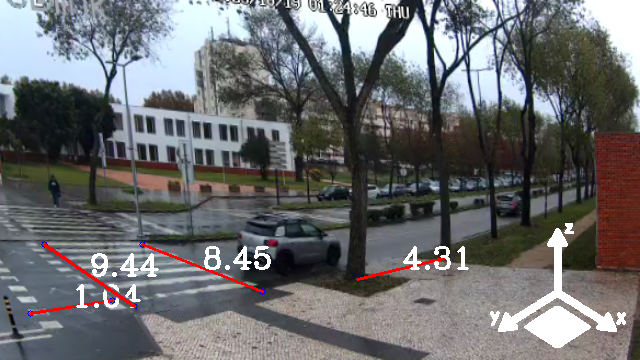}
        \caption{Measuring the crosswalk with implemented coordinate algorithm.}%
        \label{fig: verdist}%
    \end{subfigure}
    \caption{Distance between objects and distance measuring.}
\end{figure}

Figure~\ref{fig:calculation_frame} illustrates the intra-distance (red connection) and inter-distance (blue connection) between objects. It would be simpler to use a stereo-vision algorithm~\cite{stero_vision} to determine the precise location of each object, but doing so would require an increase to the implementation cost, by adding one more camera.

%Through the results of this algorithm, it is possible to have a real-time digital platform with all the information needed from the one camera~\cite{camera_coordenates}. An example of the implementation of this algorithm can be seen in Figure~\ref{fig:calculation_frame}, where, from the same system of objects, pictures from different angles are taken to ensure that the coordinate system implemented is correct~\cite{camera_coordenates}.

% \begin{figure}[h]
%   \centering
%   \includegraphics[width=0.4 \linewidth]{img/Coordenates estimate.png}
%   \caption{Example of implementation of geo-framing.}
%   \label{fig: coordenate estimate}
% \end{figure}

Now with distinct sets of data, geo-position and video camera frame analysis results for the same object, it is possible to establish a connection between the two~\cite{camera_coordenates}, in Figure~\ref{fig:calculation_frame} distances between detected object can be observed. Based on the video camera data, and real intra and inter-distances, the velocity and the vector of movement can now be calculated. Then, for each object, it is possible to predict braking distances and predict if a collision is probable to happen.

\subsection{Validating detections}

\begin{figure}[ht]
\centering
    \begin{subfigure}{0.45\linewidth}
        \centering
        \includegraphics[width=0.8\textwidth]{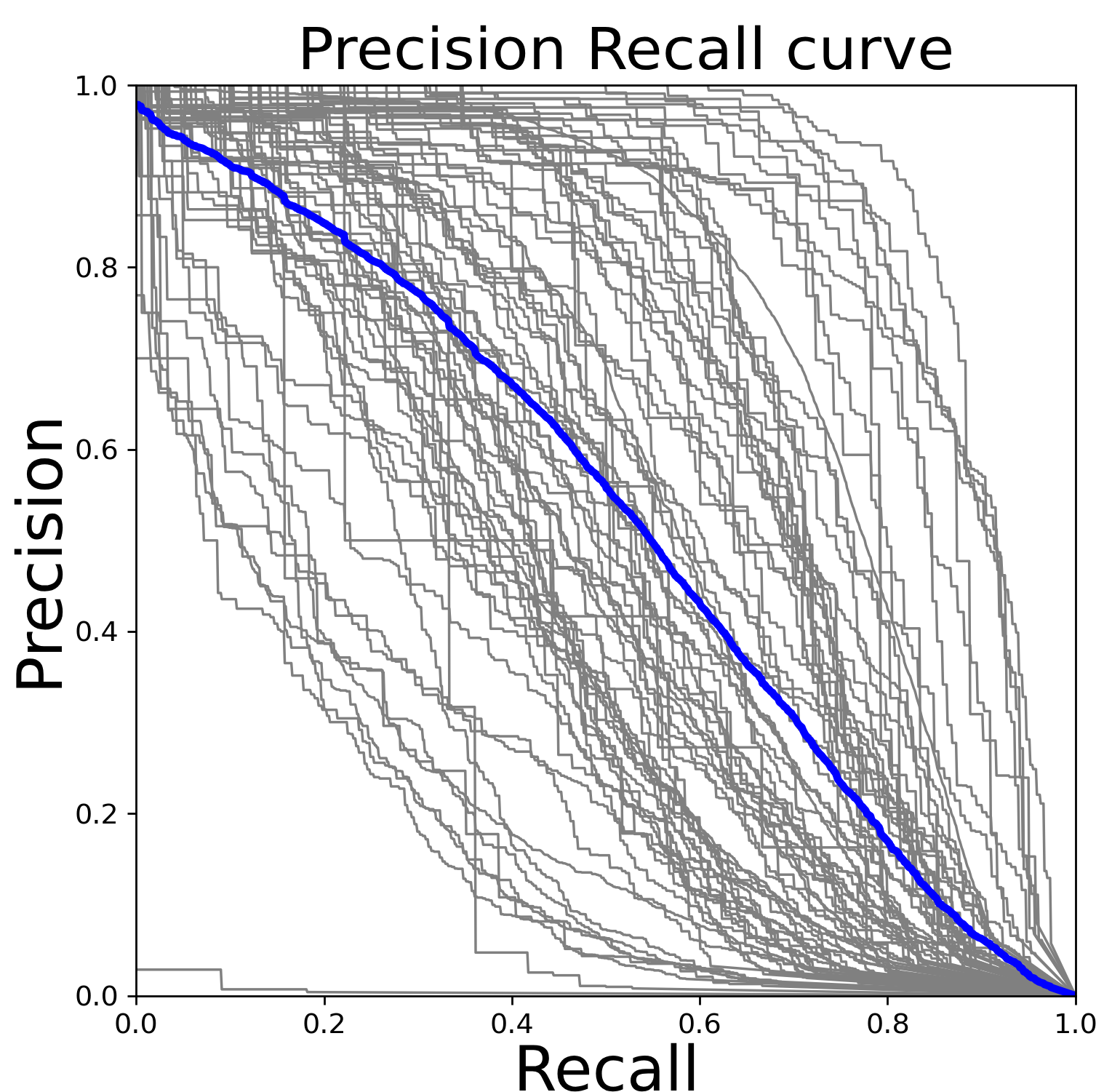}
        \caption{Precision and Recall.}
        \label{fig:Precision_and_Recall}
    \end{subfigure}
    \hfill
    \begin{subfigure}{0.45\linewidth}
        \centering
        
        \includegraphics[width=0.8\textwidth]{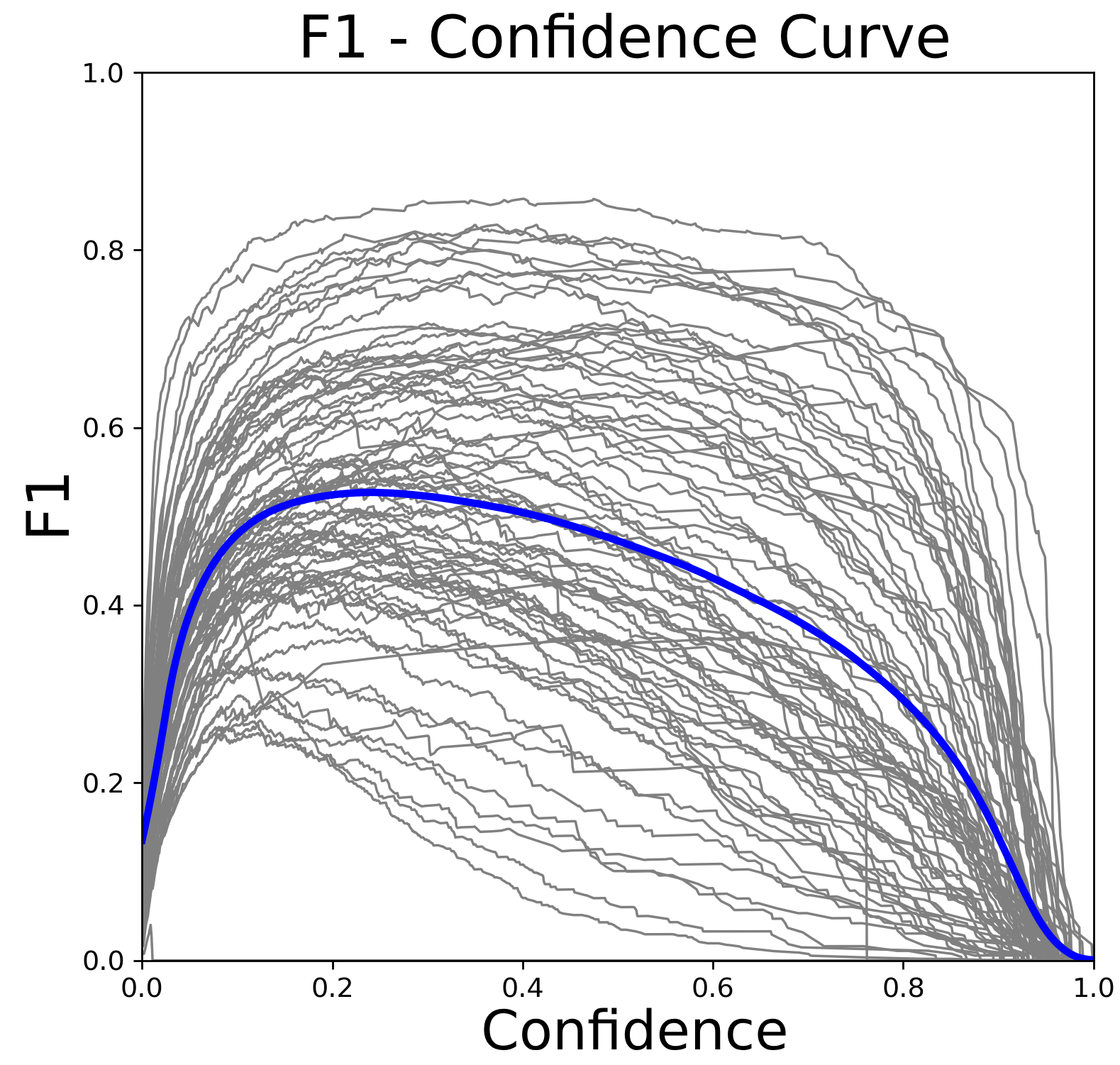}
        \caption{F1-score.}
        \label{fig:F1_test}
    \end{subfigure}
    \caption{Evaluation of the detection model.}
    \label{fig:tests}
\end{figure}

After calibrating the camera, we are able to observe an improvement in the detection confidence and, as a consequence, an improvement on the tracking. Confidence validation tests are used in \ac{yolo} to validate the object identification for accuracy and dependability. Each detected object's confidence value indicates how confident is the model on its prediction~\cite{yolov5}. To validate the model, Precision/Recall and Confidence/F1 curves were analyzed. The Precision and Recall metrics infer whether the model classifications are systematically correct and whether the set is good to generalize, the F1-score measures the overall model’s accuracy using the harmonic mean of precision and recall metrics, encouraging similar values. The results of the F1-score can be visualized in Figure~\ref{fig:tests}\subref{fig:F1_test}, and the Precision and Recall values can be found in Figure~\ref{fig:tests}\subref{fig:Precision_and_Recall}. The results show that the best value for confidence is near the $0.2$ for an optimal classification based on the trained set for the model. The confidence value of 0.2 was not used, to minimize false positives, since the algorithm was pretrained to be more generic.

\begin{figure}[h]
    \centering
    \includegraphics[width= 0.5 \linewidth]{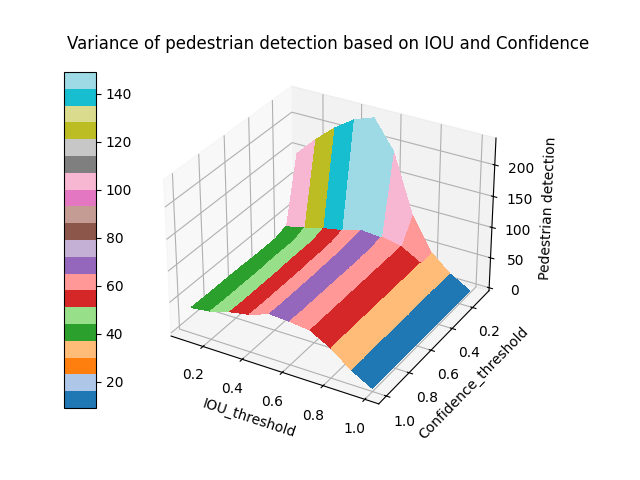}
    \caption{Confidence and intersection over union distance vs the number of detections.}%
    \label{fig: varconfiou}%
\end{figure}

The algorithm analyses each object by comparing the predicted bounding boxes confidence scores with a predetermined threshold value during validation. If the confidence score is higher than the cutoff, the detection is regarded as reliable and legitimate. If not, it is seen as a lower accuracy prediction or a false positive. As shown in Figure~\ref{fig: varconfiou}, the algorithm is more stable in mid configurations, since the \textit{intersection over union} distance also significantly affects the algorithm's ability to recognize objects. A count is performed in a prerecorded feed from the camera where it was found 66 pedestrians.

The better results are obtained for $0.6$ as the algorithm's \textit{confidence} level and the \textit{intersection over union}. It is crucial to note the algorithm is unable to conduct a high confidence identification due to the low quality of the feed and the prerecorded video, which is not in high definition. The presented values for Figure~\ref{fig: varconfiou} are as expected: if the confidence values are low, the algorithm will provide a false detection; in the intersection over union it is expected to not detect objects for high values because, as it increases the similarity between different detections, the system cannot generalize.

\subsection{Distance validation}

Distance validation for geoframing refers to the process of verifying and validating the distances between objects or locations within a geoframe. A geoframe is a defined geographic area or region of interest represented on the frame. The purpose of distance validation is to ensure that distances measured or calculated within the geoframe align with the expected or desired values.

To test how accurate the distance is set in the frame and determine reference points, using the algorithm mentioned in \cite{camera_coordenates}, it is possible to estimate and measure the distances. Taking this into consideration, measurements of the crosswalk were taken: the crosswalk measures $8$m, and the value shown in Figure~\ref{fig: verdist} is the running algorithm output, $8.45$m and $9.44$m. This confirmation is chosen because, since the geoframing validation is not done in a linear plane, it cannot take in consideration the distortion made by the Z-axis. This cannot be compared with other works, as the situation is very volatile. Moreover, related work is not suitable for comparison, since they do not evaluate this metric, as it is unique for each case.

By conducting the distance validation for geoframing, it is ensured that the accuracy and reliability of distance measurements are within a defined geographic area. This helps to maintain data integrity, and supports decision-making processes for situations within the geoframe. The considered positions for gathering the points are done in the road, so the algorithm does not recognize distances outside the considered box.

\section{Collision Detection and Risk}

For road safety, the space between objects is essential. By keeping the proper distance, cars can respond quickly to sudden movements of pedestrians, which lowers the risk of collisions. Better sight, reaction time, and mobility are made possible for both parties when \acp{vru} keep a safe distance.

Maintaining distance while driving in areas where pedestrians are present, such as near schools, residential areas, or busy downtown streets, is crucial. It gives drivers more time to react if a pedestrian suddenly steps into the road, improving the chances of avoiding accidents.

Keeping proper distances and speeds is essential for \acp{vru} safety on the road. Respecting speed limits greatly improves safety by giving vehicles more time to react and lessening the severity of potential accidents, especially in regions with a high pedestrian traffic density. This highlights how important speed regulation is to preventing collisions between vehicles and pedestrians on the roads.

\subsection{Modeling}

The risk calculation approach takes into consideration that:
\begin{itemize}
    \item \ac{vru}'s have the ability to stop in an instant, regardless of their previous velocity or acceleration;
    \item The formula serves all detected objects.
\end{itemize}

The braking distance of a vehicle at a constant velocity can be calculated using the following equation:
\begin{equation}
    d = \frac{{v^2}}{{2 \cdot \mu \cdot g}}
    \label{eq: distance}
\end{equation}

where $d$ is the braking distance in meters, $v$ is the velocity of the object in meters per second, \(\mu\) is the coefficient of friction between the tires and the road surface, and $g$ is the acceleration due to gravity, approximately $9.8 m/s^2$.

Equation \ref{eq: distance} assumes an uniform deceleration of the vehicle during braking and does not take into account factors such as driver reaction time, which is the main reason for accidents. The reaction time is an important factor, since by decreasing it, safety can be increased in the environment.

\begin{table}[h]
    \centering
    \includegraphics[width=0.5\linewidth]{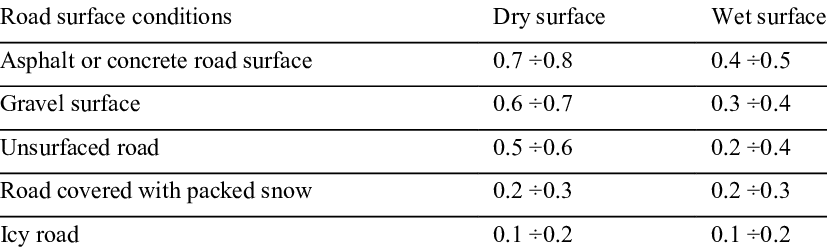}
    \caption{Friction study in roads, wet and dry, in~\cite{friction}.}
    \label{fig: friction}
\end{table}

In this work it is considered a \(\mu\) of $0.6$~\cite{friction}, as in Table \ref{fig: friction}, so that we can factor in oscillations in conditions of braking, and always prioritize safety, as the conditions in Aveiro were mostly dry during the period when tests were conducted. Practicing safe driving, maintaining a sufficient distance from the vehicle ahead, and adjusting the speed according to road and traffic conditions are essential to ensure an adequate braking distance.

\subsection{Proposed Approach}

The prediction of an accident, or a dangerous situation, is the main focus of this work. This prediction can be achieved using techniques such as geoframing, which maps points in the traffic scene to corresponding points in the projected view. The steps to calculate collision prevision and warning tickers are represented in Algorithm~\ref{algo: collision}. 

\begin{algorithm}[h]
\tiny
  \caption{Collision Detection Algorithm with YOLO}
  \begin{algorithmic}[1]
    \Require Pre-trained \ac{yolo} model, Geoframing matrix
    \State Load the pre-trained \ac{yolo} model
    \State
    \State Initialize object tracking for the two identified objects
    
    \Function{CalculateDistance}{$bbox_1, bbox_2$}
      \State Calculate the center point of $bbox_1$ and $bbox_2$ 
      \State Adjust $bbox_1$ and $bbox_2$ to the ground of object
      \State Calculate $bbox_1$ and $bbox_2$ points in real world $lat$ and $lon$, and  Haversin distance between the two center points
      \State \Return distance
    \EndFunction
    
    \State
    
    \State Start the collision detection loop
    \While{true}
      \State Capture a frame from the camera or video stream
      \State Perform object detection using the \ac{yolo} model on the frame
    \State
      \For{each detected object}
        \If{the object is one of the identified objects}
          \State Get the bounding box coordinates of the object
          \State Calculate the distance between the two objects using \textsc{CalculateDistance} function
          \State Calculate velocity
          \State Infer on colision
          \If{the collision probability is over a predefined threshold}
            \State Output a collision alert or take appropriate action
          \EndIf
        \EndIf
      \EndFor
    \EndWhile
  \end{algorithmic}
  \label{algo: collision}
\end{algorithm}

The algorithm~\ref{algo: collision} requires the YOLO model pre-loaded and the values for the geo-frame. The algorithm starts by capturing the frame and, while the frame stream is fluid, the frame is passed to the YOLO model and detection process. A detection matrix is formed with all objects classification. For every object, to establish the geographical position, an offset to the centroid, of the bounding box, is made so that it becomes aligned with the ground plane, and with the distance between two objects calculated through the CALCULATEDISTANCE function, mentioned in implementation, will provide the distance between two detections, which will be done recursively for the different objects. If the distance surpasses a defined threshold, an alert should be emitted.

\begin{figure}[ht]
    \centering
    \begin{minipage}[b]{0.45\linewidth}
        \centering
        \includegraphics[width=\linewidth]{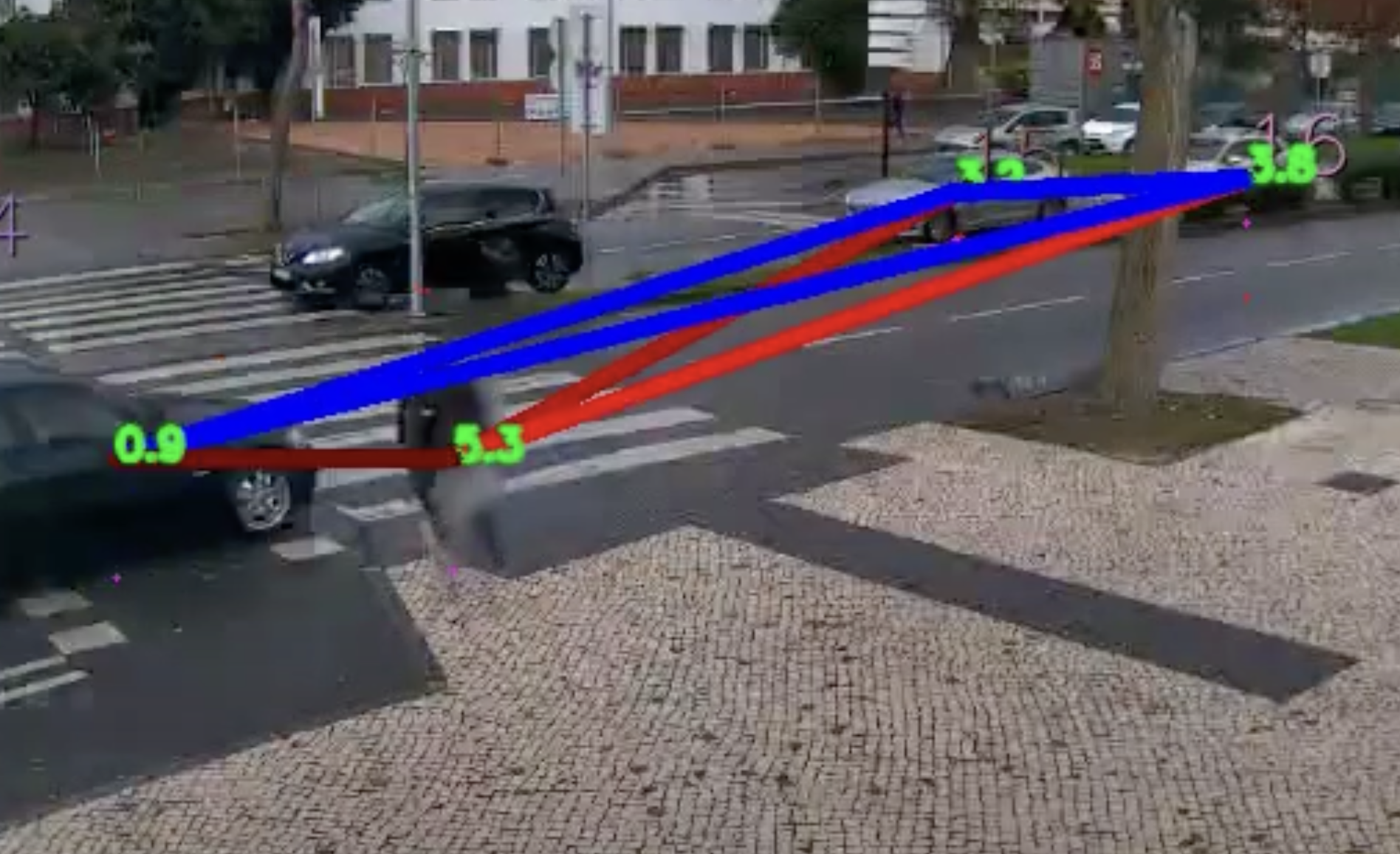}
        \caption{Algorithm processing all metrics in real-time video.}
        \label{fig: all}
    \end{minipage}
    \hspace{0cm}
    \begin{minipage}[b]{0.45\linewidth}
        \centering
        \includegraphics[width=0.7\linewidth]{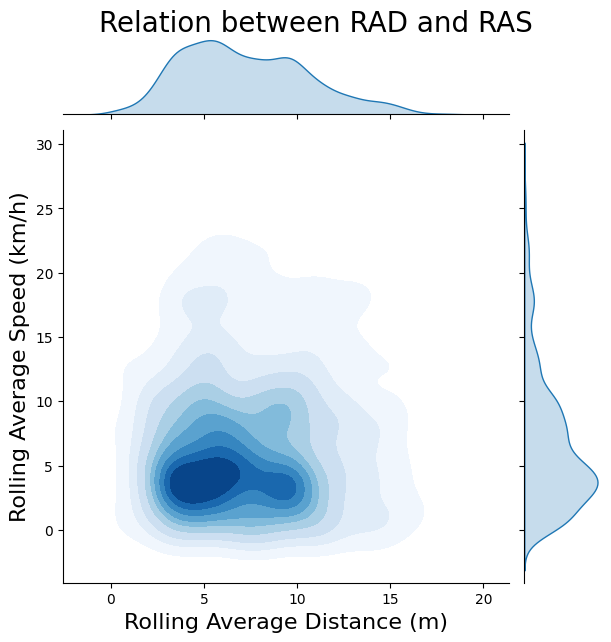}
        \caption{Relation between \ac{ras} and \ac{rad}.}
        \label{fig: RAS vs RAD}
    \end{minipage}
\end{figure}

\subsection{Results}

The distances between detected objects: those with different classification are shown in red, and if the collision is probable the red color will be brighter, while those with the same classification are showed in blue. The speed is in meters per second with the color green in the center of each object. A real-time video is available in \href{https://youtu.be/C0T9z25IlbY}{https://youtu.be/C0T9z25IlbY}.

Figure~\ref{fig: all} shows one pedestrian moving along the crosswalk and three cars aligned to cross his path. The algorithm takes in consideration the vector of movement, velocity, acceleration and position of each object. As seen in the image, the car closest to the pedestrian, has low velocity, which means it is about to stop, then collision probability is low. While the cars in the other lane (top right) have higher velocity and their movement vector is aligned to collide with the pedestrian vector, the probability of collision is higher.

%by knowing this and checking if the vector of movement has the same orientation and braking distance, is inferior to the distance of vectors colliding point, we are able to infer on an accident. 

Understanding the road state before crashes is crucial. This can be facilitated by metrics like \ac{ras} and \ac{rad}, which provide insight into the current conditions of the road. By analysing such metrics, it is not feasible to directly calculate a specific risk probability value. Instead, such as shown in Figure~\ref{fig: RAS vs RAD}, rather then giving a specific value, this graph can be divided in risk zones, and a classification can be done by the defined zones. These values are taken during a period where no accidents have occurred. This figure displays darker or more dense areas where the values are frequently seen together. These locations imply a decreased likelihood of accidents because the circumstances or actions symbolized by \ac{ras} and \ac{rad} appear to be less dangerous or unsafe when they occur frequently together.

Regions on the graph that are lighter, or more dispersed, show how infrequently these characteristics occur together. Because of the more extreme or unusual conditions or behaviors they reflect, which may result in riskier situations, these locations predict a higher possibility of accidents. Specific combinations of factors increase or decrease the likelihood of accidents (lighter regions) or decrease them (darker parts).

\section{Conclusions and Future Work}
This work focused on the development and analysis of a computer vision approach for road monitoring and safety using only a video camera in the area of monitoring. This approach is able to track objects and measure the distance of moving objects in real-time, predicting the probability of collisions between moving objects.

%The main focus of the Algorithm~\ref{algo: collision} in this context is to leverage \ac{CV} technologies to enhance crosswalk safety and pedestrian detection. \ac{CV} technologies are crucial in preventing accidents by detecting and analyzing traffic patterns and pedestrian behavior.

The results obtained from the proposed approach provide valuable insights into the performance and potential for implementation in real scenarios. To validate distances for geoframing within a specified geographic area, a geoframe was defined, by identifying reference points, measured distances, and corrected major discrepancies. Notably, elevation effects on measurements were not considered, introducing potential distortions. Nevertheless, the approach still provided reasonably accurate results, closely matching actual crosswalk distances.

Metrics played a pivotal role in assessing road conditions. Rolling averages, useful for smoothing data, capturing trends, enabling real-time assessment, supporting comparative analysis, and facilitating interpretation, were discussed. Two key metrics, \ac{ras} and \ac{rad}, inferred the road conditions based on the average vehicle and pedestrian speeds and distances, providing insights into congestion, traffic flow, and road safety. This approach stands out due to its cost-effectiveness and ease of implementation. Additionally, its innovation lies in its ability to derive metrics previously achievable only through a two-camera setup using just one camera.

Considering future work, introducing additional metrics beyond \ac{ras} and \ac{rad} would provide a more comprehensive assessment of road conditions, such as the time in use for each object. Metrics related to vehicle types, pedestrian density, or anomaly detection for unusual events on the road could be considered. Moreover, developing a user-friendly interface with visual tools for easy interaction with the monitoring system is important, since it ensures that traffic management authorities and other stakeholders can comprehend the data and analysis results effectively.

% Referências
\bibliographystyle{IEEEtran}
\bibliography{main.bib}

\end{document}